\documentclass[letterpaper, 10 pt, conference]{ieeeconf}% Use this line for a4 paper

\IEEEoverridecommandlockouts                              % This command is only needed if 
\usepackage{microtype}
\usepackage{amsmath,amssymb,amsfonts}
\usepackage{algorithmic}
\usepackage{graphicx}
\usepackage{textcomp}
\usepackage{xcolor}
\usepackage{subfig}
\usepackage{gensymb}
\usepackage{caption}
\usepackage{hyperref}
\usepackage{mwe}
\usepackage{amsfonts}
\usepackage{amssymb}
\usepackage{verbatim}
\usepackage{amsbsy}
\usepackage{siunitx}
\usepackage{tabularx, booktabs}
\usepackage{dblfloatfix}
\usepackage{cite}
\usepackage{cleveref}

\usepackage{algorithm, algorithmic}
\usepackage[autolanguage]{numprint}

\usepackage{spreadtab}
\usepackage{multirow}% http://ctan.org/pkg/multirow
\usepackage{cite}

\usepackage{hyperref}
\usepackage{xcolor}
\usepackage{algorithmic}
\usepackage{textcomp}
\usepackage{xcolor}
\usepackage{subfig}
\usepackage{gensymb}
\usepackage{caption}

\usepackage{mwe}
\usepackage{amsfonts}
\usepackage{amssymb}
\usepackage{verbatim}
\usepackage{amsbsy}
\usepackage{siunitx}
\usepackage{tabularx, booktabs}
\usepackage{soul}
\usepackage{tikz}
\usetikzlibrary{calc}

\makeatletter
\newif\if@anonymize

\@anonymizetrue    % Uncomment to hide text
\if@anonymize
  \newcommand{\highlight@DoHighlight}{
    \fill [outer sep = -15pt, inner sep = 0pt, color=black]
          ($(begin highlight)+(0,8pt)$) rectangle ($(end highlight)+(0,-3pt)$) ;
  }

  \newcommand{\highlight@BeginHighlight}{
    \coordinate (begin highlight) at (0,0) ;
  }

  \newcommand{\highlight@EndHighlight}{
    \coordinate (end highlight) at (0,0) ;
  }

  \newdimen\highlight@previous
  \newdimen\highlight@current
  \newlength{\item@width}

  \DeclareRobustCommand*\anonymize{%
    \SOUL@setup
    \def\SOUL@preamble{%
      \begin{tikzpicture}[overlay, remember picture]
        \highlight@BeginHighlight
        \highlight@EndHighlight
      \end{tikzpicture}%
    }%
    \def\SOUL@postamble{%
      \begin{tikzpicture}[overlay, remember picture]
        \highlight@EndHighlight
        \highlight@DoHighlight
      \end{tikzpicture}%
    }%
    \def\SOUL@everyhyphen{%
      \discretionary{%
        \SOUL@setkern\SOUL@hyphkern
        \SOUL@sethyphenchar
        \tikz[overlay, remember picture] \highlight@EndHighlight ;%
      }{%
      }{%
        \SOUL@setkern\SOUL@charkern
      }%
    }%
    \def\SOUL@everyexhyphen##1{%
      \SOUL@setkern\SOUL@hyphkern
      \settowidth{\item@width}{##1}%
      \makebox[\item@width]{}%
      \discretionary{%
        \tikz[overlay, remember picture] \highlight@EndHighlight ;%
      }{%
      }{%
        \SOUL@setkern\SOUL@charkern
      }%
    }%
    \def\SOUL@everysyllable{%
      \begin{tikzpicture}[overlay, remember picture]
        \path let \p0 = (begin highlight), \p1 = (0,0) in \pgfextra
          \global\highlight@previous=\y0
          \global\highlight@current =\y1
        \endpgfextra (0,0) ;
        \ifdim\highlight@current < \highlight@previous
          \highlight@DoHighlight
          \highlight@BeginHighlight
        \fi
      \end{tikzpicture}%
      \settowidth{\item@width}{\the\SOUL@syllable}%
      \makebox[\item@width]{}%
      \tikz[overlay, remember picture] \highlight@EndHighlight ;%
    }%
    \SOUL@
  }
\else
  \newcommand{\anonymize}[1]{#1}
\fi
\makeatother 

\title{\LARGE \bf
Mapless Navigation of a Hybrid Aerial Underwater Vehicle with Deep Reinforcement Learning Through Environmental Generalization}

\author{Ricardo B. Grando$^{1,2}$, Junior C. de Jesus$^{2}$, Victor A. Kich$^{3}$, Alisson H. Kolling$^{3}$, \\ Rodrigo S. Guerra$^{2}$, Paulo L. J. Drews-Jr$^{2}$
\thanks{$^{1}$Ricardo B. Grando is with the Universidad Tecnol\'ogica del Uruguay (UTEC). E-mail: {\tt\small ricardo.bedin@utec.edu.uy}}
\thanks{$^{2}$Ricardo B. Grando, Junior C. de Jesus, Rodrigo S. Guerra and P. L. J. Drews-Jr are with Centro de Ci\^encias Computacionais (C3) of  Universidade Federal do Rio Grande (FURG), RS, Brazil. E-mail: {\tt\small paulodrews@furg.br}}
\thanks{$^{3}$Victor A. Kich and Alisson H. Kolling are with the Universidade Federal de Santa Maria (UFSM), RS, Brazil. E-mail: {\tt\small victorkich@yahoo.com.br}}
}

\begin{document}

\maketitle
\thispagestyle{empty}
\pagestyle{empty}

\begin{abstract}
%REESCREVER done
%PRECISAMOS REVISAR NOVAMENTE AO FINAL DA REVISÃO
Previous works showed that Deep-RL can be applied to perform mapless navigation, including the medium transition of Hybrid Unmanned Aerial Underwater Vehicles (HUAUVs). This paper presents new approaches based on the state-of-the-art actor-critic algorithms to address the navigation and medium transition problems for a HUAUV. We show that a double critic Deep-RL with Recurrent Neural Networks improves the navigation performance of HUAUVs using solely range data and relative localization. Our Deep-RL approaches achieved better navigation and transitioning capabilities with a solid generalization of learning through distinct simulated scenarios, outperforming previous approaches.
\end{abstract}
%\vspace{-1mm}

% \begin{IEEEkeywords}
% Deep Reinforcement Learning, Hybrid Unmanned Aerial Underwater Vehicle, Mapless Navigation
% \end{IEEEkeywords}

%\vspace{-3mm}
\section*{Supplementary Material}\label{supplementary_material}

% Video of the experiments are available at: \anonymize{AAAAAAAAAAAAAAAAAAAAAAAAAAAAA}
% % \texttt{\url{https://youtu.be/rKqUMOKzgSI}}. 
% Released code at: \anonymize{
% % \texttt{\url{https://github.com/ricardoGrando/DoCRL}}. 
% AAAAAAAAAAAAAAA}

%\vspace{-2.5mm}
\section{Introduction}
\label{introduction}

% REESCREVER <= REVISEI MAS MANTIVE A ESTRUTURA E A IDEIA PRINCIPAL

Several studies about Hybrid Unmanned Aerial Underwater Vehicles (HUAUVs) have been published recently~\cite{drews2014hybrid, neto2015attitude, da2018comparative, maia2017design, lu2019multimodal, mercado2019aerial, horn20, aoki2021}. These types of vehicles enable an interesting range of new applications due to their capability to operate both in the air and underwater. These include inspection and mapping of partly submerged areas in industrial facilities, search and rescue and others. Most of the literature in the field is still focused on vehicle design, with few published works on the theme of autonomous navigation~\cite{bedin2021deep}. The ability to navigate in both environments and successfully transit from one to another imposes additional challenges that must be addressed.

Lately, approaches based on Deep-RL have been enhanced to address navigation-related tasks for a range of mobile vehicles, including ground mobile robots~\cite{ota2020efficient}, aerial robots~\cite{tong2021uav,grando2022double} and underwater robots~\cite{carlucho2018}. Based on actor-critic methods and multi-layer network structures, these approaches have achieved interesting results in mapless navigation, obstacle avoidance, even including media transitioning for HUAUVs~\cite{bedin2021deep, de2022depth}. However, the challenges faced by this kind of vehicle make these existing approaches still too limited, with poor generalization through different scenarios.

% Falar da motivação:
% Como o artigo melhora as abordagens anteriores?

\begin{figure}[tbp!]
    \vspace{-2mm}
    \centering
    \includegraphics[width=\linewidth]{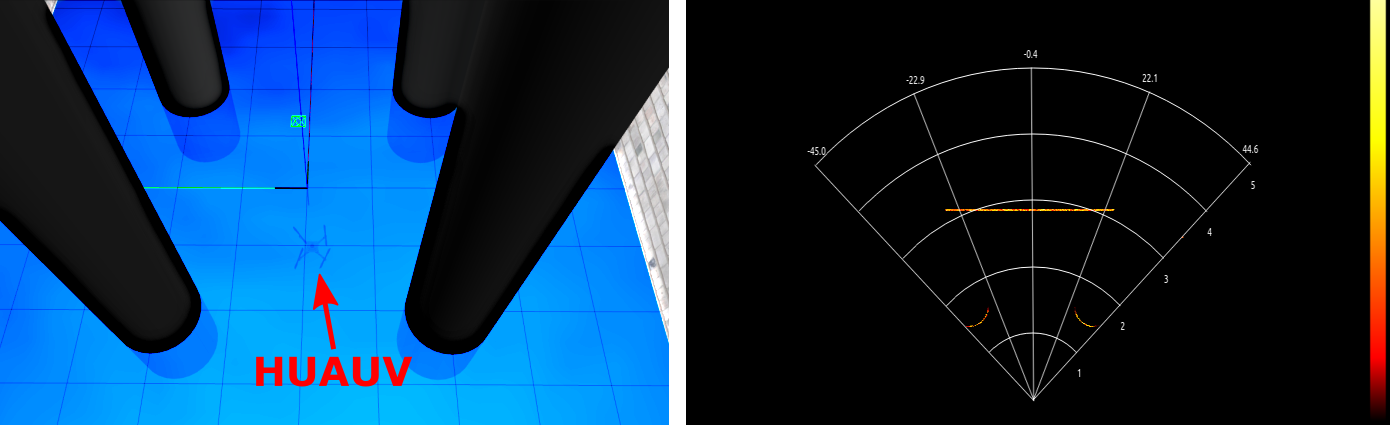}
    \caption{Our HUAUV underwater in the first scenario (left) and its respective sonar readings (right).}
    \label{fig:simenv}
    \vspace{-4mm}
\end{figure}

In this work, we present two new double-critic Deep-RL approaches in the context of HUAUVs to perform navigation-related tasks in a continuous state-space environment:  (1)~a deterministic approach based on Twin Delayed Deep Deterministic Policy Gradient (TD3)~\cite{fujimoto2018addressing}; and
 (2)~a stochastic approach based on Soft Actor-Critic (SAC)~\cite{haarnoja2018soft}. We show we are capable of training agents that are consistently better than state-of-the-art in generalizing through different simulated scenarios, with improved stability in mapless navigation, obstacle avoidance and medium transitions. Our evaluation tasks included both air-to-water and water-to-air transitions. We compared our methods with other single critic approaches and with an adapted version of a traditional Behavior-Based Algorithm (BBA)~\cite{marino2016minimalistic} used in aerial vehicles. Fig.~\ref{fig:simenv} shows a snapshot of our simulation environment.
 
%The main contributions of this work are listed as follows:

This work provides the following main contributions:

\begin{itemize}

\item We show that our agents present a robust capacity for generalization through different environments, achieving a good performance in a complex and completely unknown environment. The robot also performs the medium transition, being capable of arriving at the desired target and avoiding collisions.

\item We show that a Long Short Term Memory (LSTM) architecture can achieve better overall performance and capacity for generalization than the state-of-the-art Multi-Layer Perceptron (MLP) architectures.%, commonly used for mobile robots \cite{tai2017virtual, singh2018mobile, bedin2021deep}.

\end{itemize}

This work has the following structure: the related works are discussed in the following section (Sec. \ref{related_works}). Following it, we present our methodology in Sec. \ref{methodology}. The results are presented in Sec. \ref{results} and discussed in Sec. \ref{conclusion}. % Finally, we discuss our contributions and present future works in Sec. \ref{conclusion}. <= VAMOS FUNDIR CONCLUSIONS E DISCUSSION EM ÚNICA SEÇÃO

\section{Related Work}
\label{related_works}

% Estrutura:
% (1) Traditional / Ground
% (2) UAV (General)
% (3) UAV Mapless
% (4) HUAUV

For more traditional types of vehicles, several works have been published demonstrating how efficiently Deep-RL can solve the mapless navigation problem~\cite{tobin2017domain}. For a ground robot, Tai~\emph{et al.}~\cite{tai2017virtual} demonstrated a mapless motion planner based on the DDPG algorithm employing a 10-dimensional range finder combined with the relative distance to the target as inputs and continuous steering signals as outputs. Recently, Deep-RL methods have also been successfully used by Ota~\emph{et al.}~\cite{ota2020efficient}, de Jesus~\emph{et al.}~\cite{jesus2019deep,jesus2021soft} and others, to accomplish mapless navigation-related tasks for terrestrial mobile robots. Singh and Thongam~\cite{singh2018mobile} demonstrated efficient near-optimal navigation for a ground robot in dynamic environments employing an MLP to perform speed control while choosing collision-free path segments.

For UAVs, Kelchtermans and Tuytelaars \cite{kelchtermans2017hard} demonstrated how memory could help Deep Neural Networks (DNN) for navigation in a simulated room crossing task. Tong~\emph{et al.}~\cite{tong2021uav} showed better than state-of-the-art convergence and effectiveness in adopting a DRL-based method combined with a LSTM to navigate a UAV in highly dynamic environments, with numerous obstacles moving fast.

When it comes to problems involving specifically mapless navigation for UAVs, few works examine the effectiveness of Deep-RL. Grando~\emph{et al.}~\cite{grando2020visual} explored a Deep-RL architecture, however, navigation was constrained to a 2D space. Rodriguez \emph{et al.}~\cite{rodriguez2018deep} employed a DDPG-based strategy to solve the problem of landing UAVs on a moving platform. Similar to our work, they employed RotorS framework~\cite{furrer2016rotors} combined with the Gazebo simulator. Sampedro~\emph{et al.}~\cite{sampedro2019fully} proposed a DDPG-based strategy for search and rescue missions in indoor environments, utilizing real and simulated visual data. Kang~\emph{et al.}~\cite{kang2019generalization} also used visual information, although he focused on the subject of collision avoidance. In a go-to-target task, Barros~\emph{et al.}~\cite{2020arXiv201002293M} applied a SAC-based method for the low-level control of a UAV. Double critic-based Deep-RL approaches similar to the one proposed here have also been shown to yield good results\cite{grando2022double}.

The HUAUV literature is still mostly concerned with vehicle design and modeling \cite{drews2014hybrid, neto2015attitude, da2018comparative, maia2017design, lu2019multimodal, mercado2019aerial, horn20}. Two works have recently tackled the navigation problem with the medium transition of HUAUVs~\cite{pinheiro2021trajectory}, \cite{bedin2021deep}. Pinheiro~\emph{et al.} \cite{pinheiro2021trajectory} focused on smoothing the medium transition problem in a simulated model on MATLAB. Grando~\emph{et al.}~\cite{bedin2021deep} developed Deep-RL actor-critic approaches and a MLP architecture. These two works were developed using generic distance sensing information for aerial and underwater navigation. In contrast, our work relies on more realistic sensing data, with the simulated LIDAR and sonar being both based on real-world devices.

The HUAUV presented in this paper is based on Drews-Jr~\emph{et al.} \cite{drews2014hybrid} model, which Neto~\emph{et al.}~\cite{neto2015attitude} has largely expanded. Our work differs from the previously discussed works by only using the vehicle's relative localization data and not its explicit localization data. We also present Deep-RL approaches based on double critic techniques instead of single critic, with RNN structures instead of MLP, traditionally used for mapless navigation of mobile robots. We compare our approaches with state-of-the-art Deep-RL approaches and with a behavior-based algorithm \cite{marino2016minimalistic} adapted for hybrid vehicles to show that our new methodology improves the overall capability to generalize through distinct environments.

\section{Methodology}
\label{methodology}

In this section, we describe our simulation environment, our hybrid vehicle, and the proposed Deep-RL, detailing the network structure for both deterministic and stochastic agents. We also introduce the task that the vehicle must accomplish autonomously and the respective reward function.
            
\subsection{Deterministic Deep RL}% Reescrito por Alisson

Developing on the DQN~\cite{mnih2013playing}, Deep Deterministic Policy Gradient (DDPG)~\cite{lillicrap2015continuous} employs an actor-network where the output is a vector of real values representing the chosen action, and a second neural network to learn the target function, providing stability and making it ideal for mobile robots~\cite{jesus2019deep}. While it provides good results, DDPG still has its problems, like overestimating the Q-values, which leads to policy breaking. TD3~\cite{fujimoto2018addressing} uses DDPG as its backbone, adding some improvements, such as clipped double-Q~learning with two neural networks as targets for the Bellman error loss functions, delayed policy updates, and Gaussian noise on the target action, raising its performance.    

Our deterministic approach is based on the TD3 technique. The pseudocode can be seen in Algorithm~\ref{alg:docrl_d}.

\begin{algorithm}[!htb]
    \algsetup{linenosize=\tiny}
    \scriptsize
    \caption{Deep Reinforcement Learning Deterministic}
    \label{alg:docrl_d}
    \begin{algorithmic}[1]
        \STATE Initialize params of critic networks $\theta_{1}$, $\theta_{2}$ , and actor network $\phi$
        \STATE Initialize params of target networks $\phi^{\prime}\leftarrow\phi$, $\theta_{1}^{\prime}\leftarrow\theta_{1}$, $\theta_{2}^{\prime}\leftarrow\theta_{2}$
        \STATE Initialize replay buffer $\beta$
        \FOR{$ep = 1$ to $max\_eps$}
            \STATE reset environment state
            \FOR{$t = 0$ to $max\_steps$}
                \IF {$t < start\_steps$}
                    \STATE $a_{t} \leftarrow $ env.action\_space.sample() 
                \ELSE
                    \STATE $a_{t}\leftarrow\mu_{\phi}(s_t)+\epsilon,\ \epsilon\sim \mathcal{N}(0,OU)$
                \ENDIF
                
                \STATE $s_{t+1}$, $r_{t}$, $d_{t}$, \_ $\leftarrow$ env.step($a_{t}$)
                
                \STATE store the new transition $(s_{t}, a_{t}, r_{t}, s_{t+1}, d_{t})$ into $\beta$
                
                \IF{$t > start\_steps$}
                    \STATE Sample mini-batch $B$ of $N$ transitions $(s_{t}, a_{t}, r_{t}, s_{t+1}, d_{t})$ from $\beta$
                    
                    \STATE $a'\leftarrow\mu_{\phi^{\prime}}(s^{\prime})+\epsilon,\ \epsilon\sim clip(\mathcal{N}(0,\tilde{\sigma}), -c,\ c)$ 
                    
                    \STATE Computes target: \\ $Q_{t} \leftarrow r+\gamma*\min_{i=1,2}Q_{\theta_i}(s', a')$
                    
                    % \STATE $\mathcal{L}oss_{\theta_i}=l_{MSE}(Q_{1}, Q_{t})+l_{MSE}(Q_{2},\ Q_{t})$
                    
                    \STATE Update double critics with one step gradient descent:\\
                   $\nabla_{\theta_i} \frac{1}{N} \sum_{i \in B}(Q_t - Q_{\theta_{i}(s_{t},a_{t})})^2$  \qquad for i=1,2
                    
                    \IF {t \% $policy\_freq(t)$ == 0}
                        \STATE Update policy with one step gradient descent:\\             
                        $\nabla_{\phi}\frac{1}{N} \sum_i[\nabla_{a_{t}}Q_{\theta_{1}}(s_{t},a_{t})\vert _{a_{t}=\mu(\phi)}\nabla_{\phi}\mu_{\phi}(s_{t})]$
                        
                        Soft update for the target networks: \\
                        \STATE $\phi^{\prime}\leftarrow\tau\phi+(1-\tau)\phi^{\prime}$
                        \STATE $\theta_{i}^{\prime}\leftarrow\tau\theta_{i}+(1-\tau)\theta_{i}^{\prime}$ \qquad for i=1,2

                    \ENDIF
                \ENDIF
            \ENDFOR
        \ENDFOR
  \end{algorithmic}
\end{algorithm}

We train for $max\_steps$ steps in $max\_eps$ episodes. Our approach starts by exploring random actions for the initial $start\_steps$ steps. We use an LSTM as the actor-network $\phi$ and $\phi^{\prime}$ as its target. The double critics are also LSTM networks, denoted by $\theta_{1}$ and $\theta_{2}$, with $\theta_{1}^{\prime}$ and $\theta_{2}^{\prime}$ as their targets. The learning of both networks happens simultaneously, addressing approximation error, reducing the bias, and finding the highest Q-values. The actor target chooses the action $a^{\prime}$ based on the state $s^{\prime}$, and we add Ornstein-Uhlenbeck noise to it. The double critic targets take the tuple ($s^{\prime}$, $a^{\prime}$) and return two Q-values as outputs, from which only the minimum of the two is considered. The loss is calculated with the Mean Squared Error of the approximate value from the target networks and the value from the critic networks. We use Adaptive Moment Estimation (Adam) to minimize the loss.

We update the policy network less frequently than the value network, taking into account a $policy\_freq$ factor that increases over time by the following rule:

\begin{equation*} policy\_freq(t)=\left\lfloor\left(0.5- \frac{t}{max\_steps \times 3}\right)^{-1}\right\rfloor\end{equation*}

\subsection{Stochastic Deep RL}% Reescrito por Alisson

We also introduce a bias-stochastic actor-critic algorithm based on SAC~\cite{haarnoja2018soft}, that combines off-policy updates with a stochastic actor-critic method to learn continuous action space policies. It uses neural networks as approximation functions to learn a policy and two Q-values functions similarly to TD3. However, SAC utilizes the current stochastic policy to act without noise, providing better stability and performance, maximizing the reward and the policy's entropy, encouraging the agent to explore new states and improving training speed. We use the soft Bellman equation with neural networks as a function approximation to maximize entropy. The pseudocode can be seen in Algorithm \ref{alg:docrl_s}.

\begin{algorithm}[!htb]
    \algsetup{linenosize=\tiny}
    \scriptsize
    \caption{Deep Reinforcement Learning Stochastic}
    \label{alg:docrl_s}
    \begin{algorithmic}[1]
        \STATE Initialize params of critic networks $\theta_{1}$, $\theta_{2}$ , and actor network $\phi$
        \STATE Initialize params of target networks $\phi^{\prime}\leftarrow\phi$, $\theta_{1}^{\prime}\leftarrow\theta_{1}$, $\theta_{2}^{\prime}\leftarrow\theta_{2}$
        \STATE Initialize replay buffer $\beta$
        \FOR{$ep = 1$ to $max\_eps$}
            \STATE reset environment state
            \FOR{$t = 0$ to $max\_steps$}
                \IF {$t < start\_steps$}
                    \STATE $a_{t} \leftarrow $ env.action\_space.sample() 
                \ELSE
                    \STATE $a_t\leftarrow \text{sample from } \pi_{\phi}(\cdot|s_t)$
                \ENDIF
                
                \STATE $s_{t+1}$, $r_{t}$, $d_{t}$, \_ $\leftarrow$ env.step($a_{t}$)
                
                \STATE store the new transition $(s_{t}, a_{t}, r_{t}, s_{t+1}, d_{t})$ into $\beta$
                
                \IF{$t > start\_steps$}
                    \STATE Sample mini-batch $B$ of $N$ transitions $(s_{t}, a_{t}, r_{t}, s_{t+1}, d_{t})$ from $\beta$

                    \STATE $\tilde{a}_{t} \leftarrow \text{sample from } \pi_{\phi}(\cdot|s_t)$
                    
                    \STATE $double = ([min_{i=1,2}( Q_{\theta'_{i}}({s_{t}},{\tilde{a}_{t}}))-\alpha \log \tilde{a}_{t})])$
                    
                    \STATE $Q_t=r({s_{t}},{a_{t}})+\gamma(1-d_{t})*double$ 
                    
                    \STATE Update double critics with one step gradient descent:\\
                    $\nabla_{\theta_i} \frac{1}{N} \sum_{s_t \in B}(Q_t - Q_{\theta_{i}}(s_{t},a_{t}))^2 \text{ for } i=1,2$
                    
                    \IF {t \% $policy\_freq(t)$ == 0}
                        
                        \STATE Update policy with one step gradient descent:\\
                        $ \nabla_{\phi} \frac{1}{N} \sum_{s_t \in B} ([min_{i=1,2}( Q_{\theta_{i}}({s_{t}},{\tilde{a}_{t}}))-\alpha \log \tilde{a}_{t}])$
                        
                        \STATE Soft update for the target networks: \\
                        \STATE $\phi^{\prime}\leftarrow\tau\phi+(1-\tau)\phi^{\prime}$
                        \STATE $\theta_{i}^{\prime}\leftarrow\tau\theta_{i}+(1-\tau)\theta_{i}^{\prime}$ \qquad for i=1,2

                    \ENDIF
                \ENDIF
            \ENDFOR
        \ENDFOR
  \end{algorithmic}
\end{algorithm}

Like before, here we train for ($max\_steps$) steps in ($max\_eps$) episodes as well, exploring random actions for the first ($start\_steps$) steps. An LSTM structure was used for the policy network $\phi$. After sampling a batch $B$ from the memory $\beta$, we compute the targets for the Q-functions $Q_t({r_{t}},{s_{t+1}},{d_{t}})$, and update the Q-functions. Also, here we update the policy less frequently than the value network, using the same $policy\_freq$ factor we used in our deterministic approach. 

\subsection{Simulated Environments}% Reescrito por Alisson

Our experiments were conducted on the Gazebo simulator together with ROS, using the RotorS framework \cite{furrer2016rotors} to allow the simulation of aerial vehicles with different command levels, such as angular rates, attitude, location control and the simulation of wind with an Ornstein-Uhlenbeck noise. The underwater simulation is enabled by the UUV simulator \cite{manhaes2016uuv}, which allows the simulation of hydrostatic and hydrodynamic effects, as well as thrusters, sensors, and external perturbations. With this framework, we define the vehicle's underwater model with parameters such as the volume,  additional mass, center of buoyancy, etc., as well as the characteristics of the underwater environment itself.

\begin{figure}[tbp!]
    \vspace{-2mm}
    \centering
    \includegraphics[width=\linewidth]{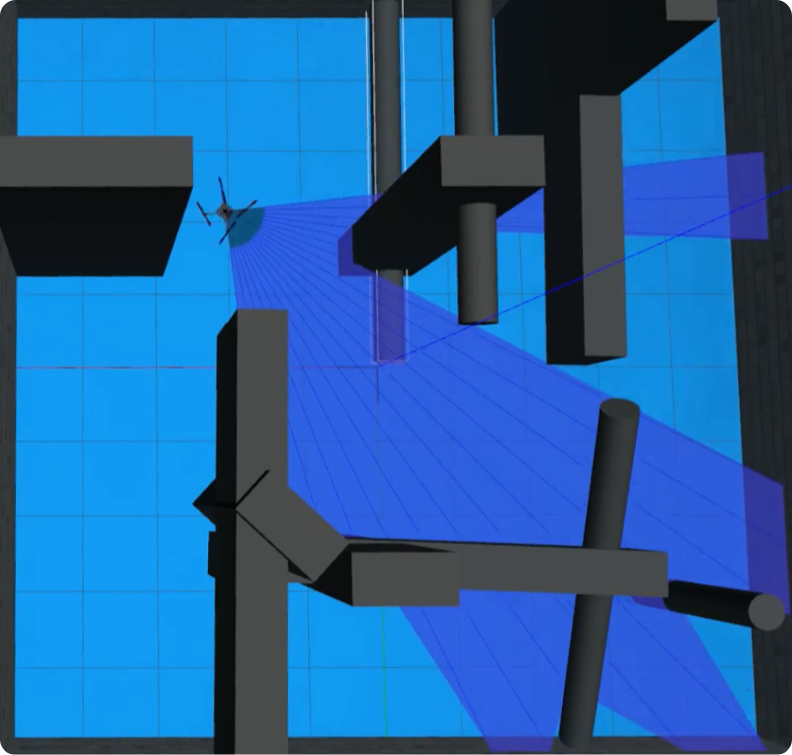}
    \caption{Our HUAUV performing in the second scenario.}
    \label{fig:env2}
    \vspace{-4mm}
\end{figure}

We developed two environments that simulate a walled water tank, with dimensions of 10$\times$10$\times$6 meters and a one-meter water column. The first environment has four cylindrical columns representing subsea drilling risers. The second environment simulates complex structures, like those found in sea platforms, and contains several elements, such as walls, half walls and pipes (Figure ~\ref{fig:env2} ).

\subsection{HUAUV Description} % Reescrito por Ricardo

% \begin{figure}[tbp!]
%     %\vspace{-1mm}
%     \centering
%     \includegraphics[width=\linewidth]{img/sonar_v3.png}
%     \caption{Our HUAUV performing in the second scenario (left) and the Sonar sensing (right).}
%     \label{fig:sonar}
%     %\vspace{-6mm}
% \end{figure}

Our vehicle was based on the model presented by Drews-Jr \emph{et al.} \cite{drews2014hybrid}, Neto \emph{et al.} \cite{neto2015attitude} and \emph{et al.} \cite{horn2019study}. We described it using its actual mechanical settings, including inertia, motor coefficients, mass, rotor velocity, and others. A ROS package containing the vehicle's description plus the Deep-RL agents can be found in the \nameref{supplementary_material}. %\footnote{\scalebox{0.92}{\url{https://github.com/ricardoGrando/hydrone\_deep\_rl\_icra}}}.

The vehicle sensing was optimized to mimic real-world LIDAR and Sonar. The described LIDAR is based on the UST 10LX model. It provides a 10 meters distance sensing with $270$\degree~of range and $0.25$\degree~of resolution, simulated using the plugin ray of Gazebo. Our simulated FLS sonar was based on the sonar simulation plugin developed by Cerqueira \emph{et al.}~\cite{cerqueira2017novel}. We described a FLS sonar with 20 meters of range, with a bin count of 1000 and a beam count of 256. The width and height angles of the beam were $90$\degree~and $15$\degree~, respectively. We obtained these values from the relative localization data using Rotors' geometric controller. In the real world, localization information can be obtained from a combination of standard localization sensing of hybrid vehicles like Global Positioning System (GPS) and Ultra Short Baseline (USBL).

%\vspace{-2mm}
\subsection{Network Structure and Rewarding System} \label{secapproach}
%
% REESCREVER

% In this work, we aim 
% % that our described HUAUV manages 
% to navigate our described HUAUV autonomously from a starting point to a target point in a different environment without any environmental knowledge, preventing any collision with the scenario by using only range readings and the vehicle's relative localization data. Its translation function, therefore, is defined as:

% \begin{equation*}
%     s_t = f(x_t, t_t, m_{t-1}),
% \end{equation*}
% %
% \noindent where $x_t$ is the range findings, $t_t$ is the target's relative information, and $m_{t-1}$ is the vehicle's motion of the last step. These variables describe the current state of $s_t$ of the vehicle. Used in many Deep-RL works for terrestrial mobile robots \cite{tai2017virtual}, \cite{jesus2019deep}, this model helps to obtain the behavior of the agent given its current state.

% \subsubsection{Networks Structure}

The structure of both our approaches has a total of 26 dimensions for the state, 20 samples for the distance sensors, the three previous actions and three values related to the target goal, which are the vehicle's relative position to the target and relative angles to the target in the x-y plane and the z-distance plane. When in the air, 20 samples come from the LIDAR. We get these samples equally spaced by $13.5\degree$ in the $270\degree$ LIDAR. When underwater, the distance information comes from the Sonar. We also get 20 beams equally spaced among the total of 256, and we take the highest bin in each beam. This conversion based on the range gives us the distance towards the obstacle or the tank's wall \cite{Santos18,Santos19}. The actions are scaled between $0$ and $0.25$ $m/s$ for the linear velocity, from $-0.25$ $m/s$ to $0.25$ $m/s$ for the altitude velocity and from $-0.25$ to $0.25$ $rad$ for the $\Delta$ yaw.

\subsubsection{Reward Function} % Reescrito por Ricardo

We proposed a binary rewarding function that yields a positive reward in case of success or a negative reward in case of failure or in case the episode ($ep$) ends at the 500 steps limit:

\vspace{-5mm}
\begin{equation}
r(s_t, a_t)= 
\begin{cases}
    r_{arrive}           & \text{if } d_t < c_d\\
    r_{collide}          & \text{if } min_x < c_o\ ||\ ep = 500\\
    % c_{r_1}(d_{t-1} - d_t)& \text{if } (d_{t-1} - d_t) \in \mathbb{R}\\
    %> 0\\
    %c_r2               & \text{if } (d_{t-1} - d_t) \leq 0\\
\end{cases}
\end{equation}

The reward $r_{arrive}$ was set to 100, while the negative reward $r_{collide}$ was set to -10. Both $c_d$ and $c_o$ distances were set to $0.5$ meters.
\section{Experimental Results}
\label{results} % Reescrito por RIcardo

In this section, the results obtained during our evaluation are shown. During the training phase, we created a randomly generated goal towards which the agent should navigate. The agents trained for a maximum of 500 steps or until they collided with an obstacle or with the tank's border. In case of reaching the goal before the limit of episodes, a new random goal was generated, allowing the total amount of reward to eventually exceed 100. A learning rate of $10^{-3}$ was used, with a minibatch of 256 samples and the Adam optimizer for all approaches, including the compared methods. We limited the number of episodes trained to 1500 episodes. The limits for the episode number ($max\_steps$) were used based on the stagnation of the maximum average reward received.

For each scenario and model, an extensive amount of statistics were collected. The task addressed is goal-oriented navigation considering medium transition, where the robot must navigate from a starting point to an endpoint. This task was addressed in two ways in our tests: (1) starting in the air, performing the medium transition and navigating to a target underwater; and the other way around, (2) starting underwater, performing the medium transition and navigating to a target in the air. We collected the statistics for each of our proposed models (Det. and Sto.) and compared them with the performance of the state-of-the-art deterministic (Det.) and stochastic (Sto.) Deep-RL methods for HUAUVs, 
% \cbox{ aqui eu chamaria mesmo de 3DNDRL-D e 3DNDRL-S e citaria o grande 2021, chama "menos atenção" para o fato do grando ser tu mesmo.}
as well as a behavior-based algorithm \cite{marino2016minimalistic} .
% \cbox{tb nao coloria tanta "atençaõ" para o BBA, chamaria de BBA, behaviour-based algorithm, e explicaria que é baseado no BBA e pronto. POis BBA ja causa uma rejeição pois é meio "toy".}
These tasks were performed for 100 trials each and we recorded the total of successful trials, the average time for both underwater ($t\_water$) and aerial ($t\_air$) navigation and their standard deviations. 

The models were all trained in the first environment and evaluated in both first (same as trained) and second (never seen) environments. We aim to outline one of the main contributions of this work, \textit{i.e.} the robust capacity to generalize of our method across environments, in this case performing in a second, unknown and more complex environment. We set the initial position for the Air-Water (A-W) trials to (0.0, 0.0, 2.5) in the Gazebo Cartesian coordinates for the two scenarios. The target position used was (3.6, -2.4, -1.0). In both environments, the target was defined in a path with obstacles on the way. Table \ref{table:mean_std} shows the results obtained for each environment for 100 navigation trials.

% We also collected the reward obtained during the training phase. In the first scenario, we collected the data for 1000 episodes, while the data of 1500 episodes are collected in the second scenario. Overall, Figures \ref{fig:reward_1} and \ref{fig:reward_2} show the rewards' moving average of the previous 300 episodes for the first and second scenarios, respectively. Table \ref{table:mean_std} shows the results obtained for each environment for 100 navigation trials.

% \begin{figure}[tbp!]
% \vspace{-4mm}
%   \subfloat[First environment.\label{fig:reward_1}]{
% 	\begin{minipage}[c][0.72\width]{
% 	   0.485\linewidth}
% 	   \centering
% 	   \includegraphics[width=\linewidth]{img/reward_env1_v3.png}
% 	\end{minipage}}
%  \hfill 	
%   \subfloat[Second environment.\label{fig:reward_2}]{
% 	\begin{minipage}[c][0.72\width]{
% 	   0.485\linewidth}
% 	   \centering
% 	   \includegraphics[width=\linewidth]{img/reward_env2_v3.png}
% 	\end{minipage}}
% \caption{Moving average of the reward over 300 episodes of training.}
% \label{fig:rewards}
% \vspace{-6.5mm}
% \end{figure}

\begin{table}[bp!]
\vspace{-5mm}
\centering
\setlength{\tabcolsep}{0.8pt}
\caption{Mean and standard deviation metrics over 100 navigation trials for all approaches in all scenarios.}
\label{table:mean_std}
\begin{tabular}{c c c c c} 
\toprule
Env & Test & $t_{air}$ (s) & $t_{water}$ (s) & Success \\
\midrule
1 & A-W Det. & $76.28$ $\pm$ $63.20$ & $12.51$ $\pm$ $20.71$ & 94 \\
1 & A-W Sto. & $21.79$ $\pm$ $4.57$ & $25.58$ $\pm$ $5.70$ & $100$ \\
% 1 & A-W 3DNDRL-LD & $13.7$ $\pm$ $1.28$ & $7.02$ $\pm$ $1.14$ & $100$ \\
% 1 & A-W 3DNDRL-2D & $12.9$ $\pm$ $0.46$ & $5.32$ $\pm$ $0.57$ & $100$ \\
1 & A-W Sto. Grando \emph{et al.} \cite{bedin2021deep} & $42.46$ $\pm$ $62.94$ & $13.13$ $\pm$ $15.15$ & $42$ \\
1 &\textbf{ A-W Det. Grando \emph{et al.} \cite{bedin2021deep}} & $\textbf{13.84}$ $\pm$ $\textbf{2.11}$ & $\textbf{5.44}$ $\pm$ $\textbf{1.73}$ & $\textbf{100}$ \\
1 & A-W BBA & $32.42$ $\pm$ $1.79$ & $21.27$ $\pm$ $0.18$ & $100$ \\
1 & W-A Det. & $24.66$ $\pm$ $10.06$ & $5.0$ $\pm$ $0.71$ & $83$ \\
1 & W-A Sto. & $79.73$ $\pm$ $27.91$ & $5.41$ $\pm$ $0.34$ & $100$ \\

2 & \textbf{A-W Det.} & $61.94$ $\pm$ $45.29$ & $\textbf{8.44}$ $\pm$ $\textbf{9.09}$ & $\textbf{73}$ \\
2 & \textbf{A-W Sto.} & $\textbf{14.89}$ $\pm$ $\textbf{1.120}$ & $18.48$ $\pm$ $6.24$ & $\textbf{94}$ \\
% 3 & A-W 3DNDRL-LD & $23.79$ $\pm$ $12.74$ & $9.67$ $\pm$ $9.49$ & $23$ \\
% 3 & A-W 3DNDRL-2D & $52.41$ $\pm$ $26.24$ & $3.43$ $\pm$ $6.21$ & $2$ \\
2 & A-W Sto. Grando \emph{et al.} \cite{bedin2021deep} & - & - & $0$ \\
2 & A-W Det. Grando \emph{et al.} \cite{bedin2021deep} & - & - & $0$ \\
2 & A-W BBA & $39.69$ $\pm$ $21.92$ & $11.32$ $\pm$ $7.46$ & $28$ \\
2 & \textbf{W-A Det.} & $\textbf{8.54}$ $\pm$ $\textbf{4.44}$ & $\textbf{4.27}$ $\pm$ $\textbf{0.47}$ & $\textbf{8}$ \\
2 & \textbf{W-A Sto.} & $\textbf{15.43}$ $\pm$ $\textbf{13.39}$ & $\textbf{6.60}$ $\pm$ $\textbf{1.75}$ & $\textbf{10}$ \\
% 3 & W-A 3DNDRL-LD & $19.05$ $\pm$ $15.53$ & $7.52$ $\pm$ $4.66$ & $11$ \\
% 3 & W-A 3DNDRL-2D & $24.96$ $\pm$ $29.69$ & $13.44$ $\pm$ $13.20$ & $6$ \\
2 & W-A Sto. Grando \emph{et al.} \cite{bedin2021deep} & - & - & $0$ \\
2 & W-A Det. Grando \emph{et al.} \cite{bedin2021deep} & - & - & $0$ \\
2 & W-A BBA & $34.3$ $\pm$ $22.93$ & $6.13$ $\pm$ $17.22$ & $8$ \\
\bottomrule
\end{tabular}
\end{table}

We also performed a complementary comparison in the second scenario. We used the models trained in the second environment to collect statistics. For a better analysis, we also performed a comparison between models in this second environment. First, we collected the data for Deterministic and Stochastic models trained only in the first environment for 1500 episodes (Env1), as shown before. Then, we trained these models for 500 more episodes in the second environment (Both). Lastly, we compared them with Deterministic and Stochastic trained only in the second environment for 1500 episodes. Table \ref{table:Comparistion_lstm} shows the obtained results.

% We also performed an analysis of the actions taken by the agents in each scenario for 100 navigation trials. Fig. \ref{fig:cmd} shows the mean of each action value and its standard deviation at each step for these trials. For the last, Fig. \ref{fig:traj_nav} shows a sample of the navigation performed by our new approaches.

\begin{table}[tp!]
\centering
\setlength{\tabcolsep}{2.5pt}
\caption{Mean and standard deviation metrics over 100 navigation trials tested in the second simulated environment, for both deterministic and stochastic models trained only in the first environment (Env1), in both first and second environments (Both), and only in the second environment (Env2).}
\label{table:Comparistion_lstm}
\begin{tabular}{c c c c c} 
\toprule
Model & $t_{air}$ (s) & $t_{water}$ (s) & Success \\
\midrule
\textbf{A-W Det. (Env1)} & $61.94$ $\pm$ $45.29$ & $\textbf{8.44}$ $\pm$ $\textbf{9.09}$ & $\textbf{73}$ \\
\textbf{A-W Sto. (Env1)} & $\textbf{14.89}$ $\pm$ $\textbf{1.120}$ & $18.48$ $\pm$ $6.24$ & $\textbf{94}$ \\
\textbf{A-W Det. (Both)} & $\textbf{14.14}$ $\pm$ $\textbf{3.77}$ & $\textbf{8.69}$ $\pm$ $\textbf{3.17}$ & $\textbf{99}$ \\
\textbf{A-W Sto. (Both)} & $16.82$ $\pm$ $2.12$ & $14.92$ $\pm$ $3.60$ & $\textbf{100}$ \\
A-W Det. (Env2) & $23.17$ $\pm$ $31.12$ & $32.53$ $\pm$ $60.28$ & $21$ \\
A-W Sto. (Env2) & $19.98$ $\pm$ $15.99$ & $49.61$ $\pm$ $27.86$ & $87$ \\

W-A Det. (Env1) & $8.54$ $\pm$ $4.44$ & $4.27$ $\pm$ $0.47$ & $8$ \\
W-A Sto. (Env1) & $15.43$ $\pm$ $13.39$ & $6.60$ $\pm$ $1.75$ & $10$ \\
\textbf{W-A Det. (Both)} & $\textbf{25.09}$ $\pm$ $\textbf{38.86}$ & $\textbf{4.62}$ $\pm$ $\textbf{0.51}$ & $\textbf{34}$ \\
\textbf{W-A Sto. (Both)} & $33.41$ $\pm$ $11.82$ & $11.40$ $\pm$ $2.38$ & $\textbf{83}$ \\
W-A Det. (Env2) & - & - & $0$ \\
W-A Sto. (Env2) & $3.73$ $\pm$ $2.97$ & $30.47$ $\pm$ $9.47$ & $1$ \\
\bottomrule
\end{tabular}
\vspace{-4mm}
\end{table}

\section{Conclusions}
\label{conclusion}

The evaluation shows an overall increase in performance in navigation through both environments. It is possible to see that our approaches achieve a consistent performance of 100 successful air-to-water navigation trials with also a consistent navigation time ($14.55 \pm 0.87$ and $11.19 \pm 2.86$). In this same scenario, the stochastic performed a little worse in air-to-water navigation but outperformed the deterministic approach in water-to-air navigation. 
In the second scenario, we can see more clearly that a double-critic-based approach with an RNN structure also has a better ability to learn and generalize the environment, including the obstacles and the medium transition. While the state-of-the-art approaches with a MLP structure were not capable of performing the task, our approaches presented once again a consistent performance, especially in air-to-water navigation. Our approaches showed an excellent ability to learn the tasks and the environmental difficulties, not only the scenario itself. That was further addressed in our additional evaluation with agents trained in the first environment only, both first and second environments and the second environment only. Overall, we can conclude that double critic approaches with recurrent neural networks present a consistent ability to learn through scenarios and environments and to generalize between them. Also, our approaches outperformed the BBA algorithm 
% \cbox{BUG2 algorithm - CHamaria de behaviour-based algorithm ou algo assim} 
in the rate of successful trials and average time in almost all situations.

It is important to mention that these approaches are extensively evaluated in a realistic simulation, including control issues and disturbances such as wind. Thus, the results indicate that our approach may achieve real-world application if the correct data from the sensing and the relative localization are correctly ensured. Finally, it is also possible to analyze that these new RNN-based approaches provided a more consistent average course of action throughout the environments.The evaluation shows an overall increase in performance in navigation through both environments. It is possible to see that our approaches achieve a consistent performance of 100 successful air-to-water navigation trials with also a consistent navigation time ($14.55 \pm 0.87$ and $11.19 \pm 2.86$). In this same scenario, the stochastic performed a little worse in air-to-water navigation but outperformed the deterministic approach in water-to-air navigation. 
In the second scenario, we can see more clearly that a double-critic-based approach with an RNN structure also has a better ability to learn and generalize the environment, including the obstacles and the medium transition. While the state-of-the-art approaches with a MLP structure were not capable of performing the task, our approaches presented once again a consistent performance, especially in air-to-water navigation. Our approaches showed an excellent ability to learn the tasks and the environmental difficulties, not only the scenario itself. That was further addressed in our additional evaluation with agents trained in the first environment only, both first and second environments and the second environment only. Overall, we can conclude that double critic approaches with recurrent neural networks present a consistent ability to learn through scenarios and environments and to generalize between them. Also, our approaches outperformed the BBA algorithm 
% \cbox{BUG2 algorithm - CHamaria de behaviour-based algorithm ou algo assim} 
in the rate of successful trials and average time in almost all situations.

It is important to mention that these approaches are extensively evaluated in a realistic simulation, including control issues and disturbances such as wind. Thus, the results indicate that our approach may achieve real-world application if the correct data from the sensing and the relative localization are correctly ensured. Finally, it is also possible to analyze that these new RNN-based approaches provided a more consistent average course of action throughout the environments.

By using physically realistic simulation in several water-tank-based scenarios, we showed that our approaches achieved an overall better capability to perform autonomous navigation, obstacle avoidance and medium transition than other approaches. Disturbances such as wind were successfully assimilated and good generalization through different scenarios was achieved. With our simple and realistic sensing approach that took into account only the range information, we presented overall better performance than the state-of-the-art and classical behavior-like algorithm. Future studies with our real HUAUV are on the way. 

%\vspace{-4mm}
\section*{Acknowledgment}

% \anonymize{xx}

%This work was partly founded by the Coordenação de Aperfeiçoamento de Pessoal de Nível Superior (CAPES) and Conselho Nacional de Desenvolvimento Científico e Tecnológico (CNPq). 
The authors would like to thank the VersusAI team. This work was partly supported by the CAPES, CNPq and PRH-ANP.

\vspace{-2mm}
\bibliographystyle{./bibliography/IEEEtran}
\bibliography{./bibliography/IEEEabrv,./bibliography/IEEEexample}

\end{document}